\documentclass{article}
\usepackage{ijcai20}

\usepackage{times}

\usepackage{url}
\usepackage[hidelinks]{hyperref}
\usepackage[utf8]{inputenc}
\usepackage[small]{caption}
\usepackage{graphicx}
\usepackage{amsmath}
\usepackage{amssymb}
\usepackage{booktabs}
\newcommand{\softmax}{\operatornamewithlimits{softmax}}

\title{Answer Generation through Unified Memories over Multiple Passages}
\author{Makoto Nakatsuji, Sohei Okui\\
NTT Resonant Inc.\\
Granparktower, 3-4-1 Shibaura,
Minato-ku, Tokyo 108-0023, Japan
\\
nakatsuji.makoto@gmail.com, okui@nttr.co.jp
}
\begin{document}
\maketitle

\begin{abstract}
Machine reading comprehension methods that generate answers by referring
to multiple passages for a question have gained much attention in AI and
NLP communities. The current methods, however, do not investigate the
relationships among multiple passages in the answer generation process,
even though topics correlated among the passages may be answer
candidates. Our method, called neural answer Generation through Unified
Memories over Multiple Passages (GUM-MP), solves this problem as
follows. First, it determines which tokens in the passages are matched
to the question. In particular, it investigates matches between tokens
in positive passages, which are assigned to the question, and those in
negative passages, which are not related to the question. Next, it
determines which tokens in the passage are matched to other passages
assigned to the same question and at the same time it investigates the
topics in which they are matched. Finally, it encodes the token
sequences with the above two matching results into unified memories in
the passage encoders and learns the answer sequence by using an
encoder-decoder with a multiple-pointer-generator mechanism. As a
result, GUM-MP can generate answers by pointing to important tokens
present across passages. Evaluations indicate that GUM-MP generates much
more accurate results than the current models do.
\end{abstract}

\section{Introduction}
Machine Reading Comprehension (MRC) methods
\cite{DBLP:conf/nips/NguyenRSGTMD16,DBLP:journals/corr/RajpurkarZLL16}
that empower computers with the ability to read and comprehend knowledge
and then answer questions from textual data have made rapid progress in
recent years. Most methods try to answer a question by extracting exact
text spans from the passages retrieved by search engines
\cite{wang-etal-2018-multi-passage}, while a few try to generate answers
by copying word tokens from passages in decoding answer tokens
\cite{DBLP:conf/naacl/SongWHZG18,DBLP:journals/corr/abs-1901-02262}. The
passages are usually related to the question; thus, the descriptions
among the passages are often related to each other. The current methods,
however, do not analyze the relationships among passages in the answer
generation process. Thus, they may become confused by several different
but related answer descriptions or unrelated descriptions in multiple
passages assigned to the question. This lowers the accuracy of the
generated answers
\cite{jia-liang-2017-adversarial,wang-etal-2018-multi-passage}.

Table \ref{tab:wfaEx} lists examples from the MS-MARCO dataset
\cite{DBLP:conf/nips/NguyenRSGTMD16}, which we used in our evaluation,
to explain the problem. This table contains the question, passages
prepared for this question, and the answer given by human editors. The
phrases in bold font include the answer candidates to the question
``what is the largest spider in the world?''. They are described across
multiple passages, and some describe different spiders. There are also
descriptions that are unrelated to the question, for example, about how
humans feel about spiders or the characteristics of spiders. The
presence of several different answer descriptions or unrelated ones tend
to confuse the current answer-generation methods, and this lowers their
accuracy.

{\tabcolsep=1.7mm
 \begin{table*}[t]
 \begin{center}

\doublerulesep=1mm \caption{Example entry in the MS-MARCO dataset.}
 \footnotesize
{\tabcolsep = 1.0mm
 \begin{tabular}{c|p{16.0cm}} \hline
Question & What is the largest spider in the world? \\
\hline
Passage 1 & Top 10 largest spiders in the world! Some people scare usual
 spiders to death, while some find these little pests pretty
 harmless and not disgusting at all. but there are some monsters
 that may give creeps even to the bravest and the most skeptical. 
\\ \hline 
Passage 2 & According to the guinness book of world records, {\bf the
 world’s largest spider is the goliath birdeater native to south
 america}. Scientists say the world's largest spider, the goliath
 birdeater, can grow to be the size of a puppy and have legs
 spanning up to a foot, according to video from geobeats.
 \\ \hline
Passage 3 & {\bf The giant huntsman spider is a species of huntsman spider, a family of large}, fast spiders that actively hunt down prey. {\bf It is considered the world's largest spider by leg span}, which can reach up to 1 foot ( 30 centimeters ). \\ \hline
Answer & The giant huntsman is the largest spider in the world.\\ \hline
\end{tabular} }
 \normalsize
 \label{tab:wfaEx} \vspace{-4mm}\end{center}
\end{table*} }

To solve this problem, we propose neural answer Generation through
Unified Memories over Multiple Passages (GUM-MP). This model has
question and passage encoders, Multi-Perspective Memories (MPMs)
\cite{DBLP:conf/naacl/SongWHZG18}, Unified Memories (UMs), and an answer
decoder with a multiple-pointer-generator mechanism (see
Fig. \ref{fig:1}). It is founded upon two main ideas:

(1) GUM-MP learns which tokens in the passages are truly important for
the question by utilizing positive passages that are prepared for the
question and negative passages that are not related to the question. In
the passage encoders, it receives a question and positive/negative
passages on which it performs passage understanding by matching the
question embedding with each token embedded in the passages from
multiple perspectives. Then it encodes those information into MPMs. In
particular, it investigates the difference between matches computed for
positive passages and those for negative passages to determine the
important tokens in the passages. This avoids confusion caused by
descriptions that are not directly related to the question. For example,
a phrase like ``according to the guinness book of world records'' in
Table \ref{tab:wfaEx} can appear in passages that answer different
questions (i.e. negative passages for the current question). GUM-MP can
filter out the tokens in this phrase in generating an answer.

(2) GUM-MP computes the match between each token embedding in MPM for
each passage and the embedding composed from the rest of passages
assigned to the question. First, it picks up the target passage. Next,
it encodes the sequences of embeddings in the rest of passages into a
fixed-dimensional latent semantic space, Passages Alignment Memory
(PAM). PAM thus holds the semantically related or unrelated topics
described in the passages together with the question context. Then, it
computes the match between each embedding in the target MPM and the
PAM. Finally, it encodes the embedding sequence in the target MPM with
the above matching results into the UM. GUM-MP builds UMs for all
passages by changing the target passage. As a result, for example, in
Table \ref{tab:wfaEx}, GUM-MP can distinguish the topics of large
spiders, that of human feelings, and that of the spider characteristics
by referring to UMs in generating an answer.

Finally, GUM-MP computes the vocabulary and attention distributions for
multiple passages by applying encoder-decoder with a
multiple-pointer-generator mechanism, wherein the ordinary
pointer-generator mechanism \cite{P17-1099} is extended to handle
multiple passages. As a result, GUM-MP can generate answers by pointing
to different descriptions across multiple passages and comprehensively
assess which tokens are important or not.

We used the MS-MARCO dataset and a community-QA dataset of a Japanese QA
service, Oshiete goo, in our evaluations since they provide answers with
multiple passages assigned to questions. The results show that GUM-MP
outperforms existing state-of-the-art methods of answer generation.

\section{Related work}
\label{sec:related} Most MRC methods aim to answer a question with exact
text spans taken from evidence passages
\cite{wei2018fast,DBLP:journals/corr/RajpurkarZLL16,DBLP:conf/emnlp/YangYM15,DBLP:journals/corr/JoshiCWZ17}. Several
studies on the MS-MARCO dataset
\cite{DBLP:conf/nips/NguyenRSGTMD16,DBLP:conf/naacl/SongWHZG18,DBLP:conf/aaai/TanWYDLZ18}
define the task as answering a question using information from multiple
passages. Among them, S-Net \cite{DBLP:conf/aaai/TanWYDLZ18} developed
an extraction-then-synthesis framework to synthesize answers from the
extracted results. MPQG \cite{DBLP:conf/naacl/SongWHZG18} performs
question understanding by matching the question with a passage from
multiple perspectives and encodes the matching results into MPM. It then
generates answers by using an attention-based LSTM with a
pointer-generator mechanism. However, it can not handle multiple
passages for a question or investigate the relationships among passages
when it generates an answer. Several models based on Transformer
\cite{vaswani2017attention} or BERT
\cite{DBLP:journals/corr/abs-1810-04805} have recently been proposed in
the MRC area
\cite{DBLP:journals/corr/abs-1904-08375,DBLP:journals/corr/abs-1804-07888,article,DBLP:journals/corr/abs-1808-05759}. In
particular, \cite{DBLP:journals/corr/abs-1901-02262} is for generating
answers on the MS-MARCO dataset. These methods, however, do not utilize
positive/negative passages to examine which word tokens are important or
analyze relationships among passages, to improve their accuracy.

Regarding studies that compute the mutual attention among documents,
\cite{hao-etal-2017-end} examined cross attention between the question
and the answer. Co-attention models
\cite{DBLP:conf/iclr/XiongZS17,zhong2018coarsegrain} as well use
co-dependent representations of the question and the passage in order to
focus on relevant parts of both. They, however, do not compute the
mutual attentions among passages and only focus on the attentions
between question and passages. V-Net \cite{wang-etal-2018-multi-passage}
extracts text spans as answer candidates from passages and then verifies
whether they are related or not from their content representations. It
selects the answer from among the candidates, but does not generate
answers.

The neural answer selection method \cite{TanSXZ16} achieves high
selection accuracy by improving the matching strategy through the use of
positive and negative answers for the questions, where the negative
answers are randomly chosen from the entire answer space except for the
positive answers. There are, however, no answer-generation methods that
utilize negative answers to improve answer accuracy.

\section{Preliminary} 
\label{sec:Preliminary} Here, we explain the encoding mechanism of MPM used in MPQG, since we base our ideas on its framework.

The model takes two components as input: a passage and a question. The
passage is a sequence ${\bf{P}}= ({\bf{p}}_1, \ldots, {\bf{p}}_i,
\ldots, {\bf{p}}_{N_{p}})$ of length $N_p$, and the question is a
sequence ${\bf{Q}}=({\bf{q}}_1 , \ldots, {\bf{q}}_i , \ldots,
{\bf{q}}_{N_q})$ of length $N_q$. The model generates the output
sequence ${\bf{X}} = ({\bf{x}}_1, \ldots, {\bf{x}}_i , \ldots,
{\bf{x}}_{N_x})$ of length $N_x$ word by word. Here, ${\bf{p}}_i$ (or
${\bf{q}}_i$, ${\bf{x}}_i$) denotes a one-of-$D$ embedding of the $i$-th
word in a sequence ${\bf{P}}$ (or ${\bf{Q}}$, ${\bf{X}}$) of length
$N_p$ (or $N_q$, $N_x$).

The model follows the encoder-decoder framework. The encoder matches
each time step of the passage against all time steps of the question
from multiple perspectives and encodes the matching result into the
MPM. The decoder generates the output sequence one word at a time based
on the MPM. It is almost the same as the normal pointer-generator
mechanism \cite{P17-1099}; thus, we will omit its explanation.

MPQG uses a BiLSTM encoder, which encodes the question in both
directions to better capture the overall meaning of the question. It
processes in both directions, $\{{\bf{q}}_1, \cdots , {\bf{q}}_{N_q}\}$
and $\{{\bf{q}}_{N_q}, \cdots , {\bf{q}}_{1}\}$, sequentially. At time
step $t$, the encoder updates the hidden state by ${\bf{h}}^q_t\!\!
=\!\! [{\bf{h}}^{q,f}_t, {\bf{h}}^{q,b}_t]^{\mathrm{T}}$, where
${\bf{h}}^{q,f}_t\!\! =\!\! f({\bf{q}}_{t-1}, {\bf{h}}^{q,f}_{t-1})$ and
${\bf{h}}^{q,b}_t\!\! =\!\! f({\bf{q}}_{t+1},
{\bf{h}}^{q,b}_{t+1})$. $f()$ is an LSTM unit. ${\bf{h}}^{q,f}_t$ and
${\bf{h}}^{q,b}_t$ are hidden states output by the forward LSTM and
backward LSTM, respectively. MPQG then applies a max-pooling layer to
all hidden vectors yielded by the question sequence to extract the most
salient signal for each word. As a result, it generates a fixed-sized
distributed vector representation of the question, ${\bf{o}}^q$.

Next, MPQG computes the matching vector ${\bf{m}}_i$ by using a function
$f_m$ to match two vectors, ${\bf{o}}^q$ and each forward (or backward)
hidden vector, ${\bf{h}}^{p,f}$ (or ${\bf{h}}^{p,b}$), output from the
passage. In particular, it uses a multi-perspective cosine matching
function defined as: ${{m}}_{i,z} = f_m({\bf{h}}^{p,f}_i, {{\bf{o}}^{q}}
; {\bf{w}}_z) =\cos( {\bf{h}}^{p,f}_i \circ {\bf{w}}_z , {\bf{o}}^q
\circ {\bf{w}}_z )$, where the matrix ${\bf{W}} \in \mathbb{R}^{Z \times D}$ is a
learnable parameter of a multi-perspective weight, $Z$ is the number of
perspectives, the $z$-th row vector ${{\bf{w}}_{z}} \in {\bf{W}}$
represents the weighting vector associated with the $z$-th perspective,
and $\circ$ is the element-wise multiplication operation. The final
matching vector for each time step of the passage is updated by
concatenating the matching results of the forward and backward
operations. MPQG employs another BiLSTM layer on top of the matching
layer to smooth the matching results. Finally, it concatenates the
hidden vector of the passage ${\bf{h}}_{i}^{p}$ and matching vector
${{\bf{m}}_i}$ to form the hidden vector ${\bf{h}}_{i} \in {\bf{H}}$
(the length is $2(D\!+\!Z)$) in the MPM of the passage, which
contains both the passage information and matching information. The MPM
for the question is encoded in the same way.

\begin{figure*}[t]
\begin{center}
\includegraphics[width=\linewidth]{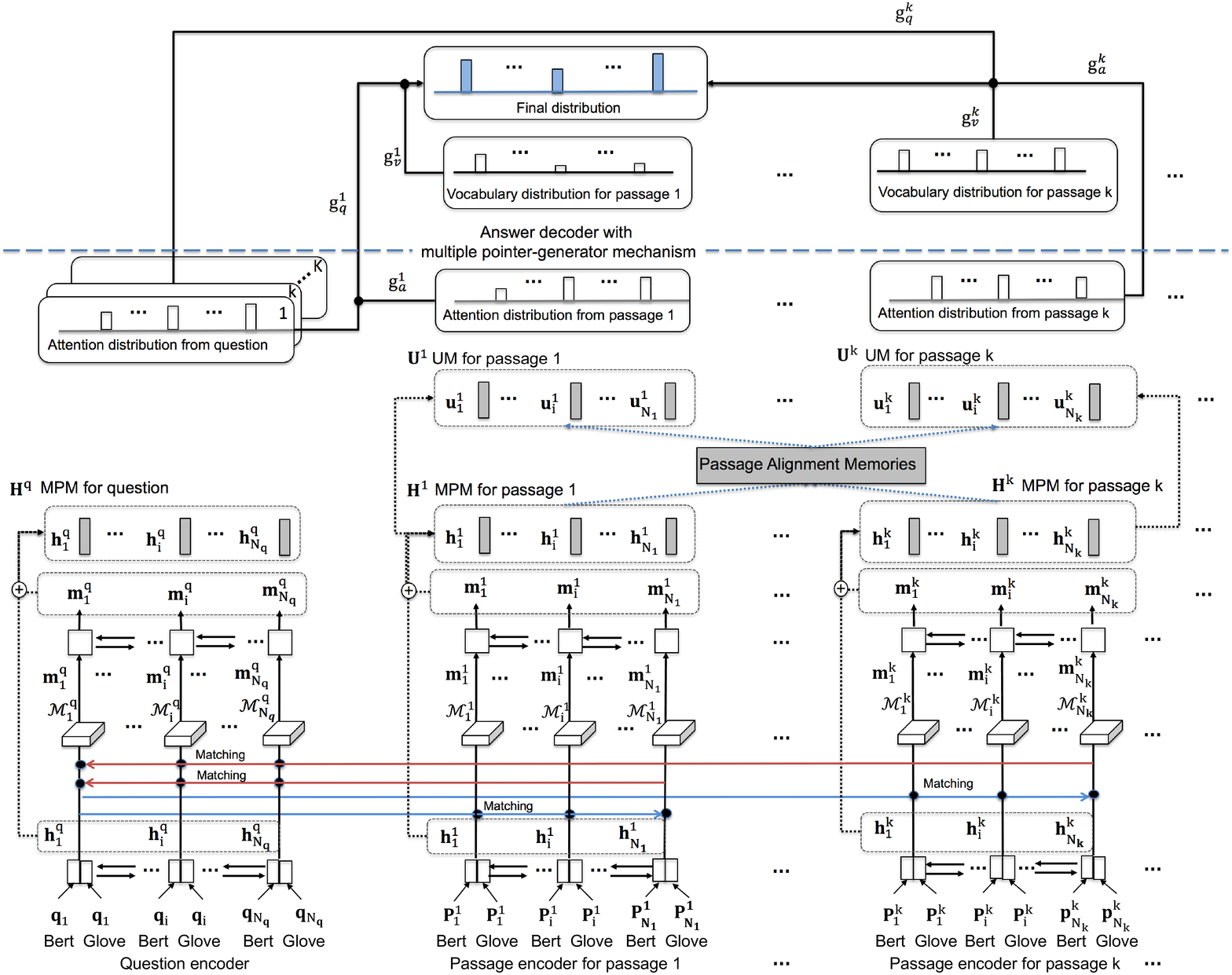}
\end{center}
 \vspace{-3mm}
\caption{Overview of GUM-MP.}
\label{fig:1}
\vspace{-3mm}
\end{figure*}

\section{Model}
Given an input question sequence ${\bf{Q}}$ and $K$ passage sequences
(e.g., the $k$-th passage is denoted as ${\bf{P}}^k = \{{\bf{p}}^k_{1},
\cdots , {\bf{p}}^k_{i}, \cdots , {\bf{p}}^k_{N_k}\}$), GUM-MP outputs
an answer sequence ${\bf{A}}= \{{\bf{a}}_1, \cdots , {\bf{a}}_t, \cdots
, {\bf{a}}_{N_a}\}$. Please see Fig. \ref{fig:1} also.

\subsection{Passage and question encoders}
The passage encoder takes a set of passage sequences $\{{\bf{P}}^1,
\cdots , {\bf{P}}^k, \cdots ,{\bf{P}}^K\}$ and a question sequence
${\bf{Q}}$ as inputs and encodes the MPMs of the passages (e.g. the MPM
of the $k$-th passage is denoted as ${\bf{H}}^{k}= \{{\bf{h}}^{k}_{1},
\cdots , {\bf{h}}^{k}_{i}, \cdots , {\bf{h}}^{k}_{{N_k}}\}$). Word
embeddings in inputs are concatenations of Glove embeddings
\cite{Pennington14glove:global} with Bert ones. The computation of
${\bf{H}}^{k}$ follows the MPQG approach explained in the previous
section; however, GUM-MP improves on MPQG by computing the match in more
detail by introducing a matching tensor ${\cal{M}}^k$ for passage
$k$. Let us explain how to compute the matching tensor. Each entry in
${\cal{M}}^{k}$ stores the matching score $m^k_{i,j,z}$ in the $z$-th
perspective between the question vector ${\bf{o}}^q$ and the $i$-th
hidden vector ${\bf{h}}^{k}_i$ (we regard passages prepared to the
question to be positive passages) and the $j$-th hidden vector
${\bf{h}}^{k,-}_j$ for negative passages (there is one negative passage
for each positive one randomly chosen from passages, which are not
assigned to the current question, in the training dataset). It is
computed as follows:
\begin{eqnarray}
 m_{i,j,z}^{k} \!\!=\!\! f_m({\bf{h}}^{k}_i, \!{{\bf{o}}^{q}};
 {\bf{w}}_z\!) \!\!-\!\! f_m( {\bf{h}}^{k,-}_j\!\!, {{\bf{o}}^{q}}; {\bf{w}}_z\!). \label{eq:mijk} 
\end{eqnarray}

GUM-MP then computes the $i$-th $Z$-dimensional matching vector ${\bf{m}}^{k}_{i}$ in ${\bf{H}}^{k}$ as follows:
 \vspace{-3mm}

\begin{eqnarray}
 {\bf{m}}^{k}_{i}\!\!\!\!\! &\!\!=\!\!&\!\!\!\!\! \sum_{j=1}^{N_k^{-}} (\alpha^{k}_{i,j} {\bf{m}}_{i,j}^{k}) \quad \mbox{ s.t. }\nonumber \\ 
\alpha^{k}_{i,j}\!\!\!\!\! &\!\!=\!\!&\!\!\!\!\! \frac{\exp(e^{k}_{i,j})}{\sum_{j=1}^{N_k^{-}}\! \exp(e^{k}_{i,j})},
 e^{k}_{i,j}\!\! =\!\! 
\tanh ( \!{\bf{w}}^{+} \!\! \cdot\! {\bf{o}}^{k}\! \! +\!
 {\bf{w}}^{-}\!\!\! \cdot\! {\bf{o}}^{k,-}\! \!\! +\! {\bf{w}}^m \! \!\cdot \! {\bf{m}}_{i,j}^k)
 \nonumber 
\end{eqnarray}
where ${\bf{w}}^{+}$, ${\bf{w}}^{-}$, and ${\bf{w}}^{m}$ are learnable
parameters. ${N_k^{-}}$ denotes the length of the negative passages
prepared for the $k$-th passage. ${\bf{m}}_{i,j}^{k}$ is a
$Z$-dimensional matching vector whose elements are multi-perspective
matching scores with the $i$-th token in the positive passage $k$ and
the $j$-th token in the negative passage (see
Eq. (\ref{eq:mijk})). ${\bf{o}}^k$ and ${\bf{o}}^{k,-}$ are positive and
negative passage vectors, respectively, computed in the same way as
${{\bf{o}}^{q}}$. The computed matching vector ${\bf{m}}^{k}_{i}$
considers the margin difference between positive and negative passages
for the $i$-th token in the $k$-th passage and thus is more concise than
the matching vector yielded by MPQG.

The question encoder also encodes the MPM of the question, ${\bf{H}}^q=
\{{\bf{h}}^q_1, \cdots , {\bf{h}}^q_i, \cdots , {\bf{h}}^q_{N_q}\}$. The
computation of ${\bf{H}}^q$ follows the computation of ${\bf{H}}^k$
except that it switches the roles of the question and passages. One
difference is that there are $K$ ${\bf{H}}^{q,k}$s, since there are $K$
passages. GUM-MP averages those ${\bf{H}}^{q,k}$s to compute a single
MPM for the question, ${\bf{H}}^q$ and thereby reduce the complexity of
the computation.

\subsection{Unified Memories}
\label{sec:um-pam} GUM-MP computes the correlations between each hidden
vector in the passage and the embedding of the rest of passages. This is
because correlated topics among passages tend to include important
topics for the question and thus may be possible answers.

First, GUM-MP picks up the $i$-th MPM. Next, it encodes the sequences of
hidden vectors in the rest of MPMs (i.e. $\{{\bf{H}}^1, \cdots ,
{\bf{H}}^{i-1}, {\bf{H}}^{i+1}, \cdots {\bf{H}}^K\}$, whose size is
$\sum_{k;k \ne i}^K{N_k}\!\! \times \!\! 2(D\!+\!Z)$), into $2(D\!+\!Z)\!
\times \! L$
latent semantics, i.e. Passage Alignment Memory (PAM) for the $i$-th
passage (we denote this as ${\bf{PA}}^i$). We say that this memory is
``aligned'' since the hidden vectors in the rest of the MPMs are aligned
through the shared weighting matrix ${\bf{W}}^p$ (the size is $\sum_{k;
k \ne i}^K{N_k}\! \times \! L$) as:

\vspace{-3mm}
\begin{equation}
 {\bf{PA}}^{i} = \{{\bf{H}}^1, \cdots , {\bf{H}}^{i-1},
 {\bf{H}}^{i+1}, \cdots , {\bf{H}}^K\}^{T} \cdot {\bf{W}}^p.\nonumber
\end{equation}
Then, GUM-MP computes the UM that unifies the information of the token
embeddings with the matching results from the question context as well
as those about the topics described among the passages. It computes the
$j$-th hidden vector in the $i$-th UM, ${\bf{u}}^{i}_j$, by
concatenating the $j$-th hidden vector in the $i$-th MPM,
${\bf{h}}^{i}_j$, with the inner product of ${\bf{h}}^{i}_j$ and
${\bf{PA}}^i$ as: ${{\bf{u}}}^{i}_{j} = [{{{\bf{h}}}^{i}_{j}},
\hspace{1mm} ({{{\bf{h}}}^{i}_{j} \cdot {{\bf{PA}}}^i})]$.  The length
is $2(D\!+\!Z)\!+\!L$.  Here, the inner product of ${\bf{h}}^{i}_j$ and
${\bf{PA}}^i$ includes information on which tokens in the passage are
matched to the other passages assigned to the same question.

Thus, GUM-MP can point the important tokens considering the correlated
topics among passages stored in UM in the decoder, which we will
describe next.

\subsection{Decoder}
The decoder is based on attention-based LSTMs with a multiple-pointer-generator mechanism.

Our multiple-pointer-generator mechanism generates the vocabulary
distribution, attention distribution for the question, and attention
distribution for the passage independently for each passage. It then
aligns these distributions across multiple passages (see
Eq. (\ref{eq:final}) described later). This is different from the
approach that first concatenates the passages into a single merged
passage and then applies the pointer-generator mechanism to the merged
passage \cite{DBLP:journals/corr/abs-1901-02262}.

GUM-MP requires six different inputs for generating the $t$-th answer
word $a_t$: (1) $K$ UMs for passages (e.g., the $k$-th UM is denoted as
${{\bf{U}}}^{k} = \{{{\bf{u}}}^{k}_{1} , .. , {{\bf{u}}}^{k}_{i} , .. ,
{{\bf{u}}}^{k}_{N_k} \}$, where each vector ${{\bf{u}}}^{k}_{i} \in
{{\bf{U}}}^{k}$ is aligned with the $i$-th word in the $k$-th passage);
(2) the MPM for the question, ${{\bf{H}}}^q = \{{{\bf{h}}}^q_1 , .. ,
{{\bf{h}}}^q_i , .. , {{\bf{h}}}^q_{N_q} \}$, where each vector
${{\bf{h}}}^q_i \in {{\bf{H}}}^q$ is aligned with the $i$-th word in the
question; (3) $K$ previous hidden states of the LSTM model,
${{\bf{s}}}^k_{t-1}$; (4) the embedding of the previously generated
word, ${{\bf{a}}}_{t-1}$; (5) $K$ previous context vectors,
e.g. ${{\bf{c}}}^{k}_{t-1}$, which are computed using the attention
mechanism, with ${{\bf{U}}}^{k}$ being the attentional memory; (6) $K$
previous context vectors, ${{\bf{c}}}^{q,k}_{t-1}$, which are computed
using the attention mechanism, with ${{\bf{H}}}^{q}$ being the
attentional memory. At $t = 1$, we initialize ${{\bf{s}}}_{0}$,
${{\bf{c}}}^{k}_{0}$, and ${{\bf{c}}}^{q,k}_{0}$ as zero vectors and set
${{\bf{a}}}_{0}$ to be the embedding of the token ``\verb|<|s\verb|>|''.

For each time step $t$ and each passage $k$, the decoder first feeds the
concatenation of the previous word embedding, ${\bf{a}}_{t-1}$, and
context vectors, ${{\bf{c}}}^{k}_{t-1}$ and ${{\bf{c}}}^{q,k}_{t-1}$,
into the LSTM model to update the hidden state: ${{\bf{s}}}^k_t =
f({{\bf{s}}}^k_{t-1}, [{{\bf{a}}}_{t-1},
{{\bf{c}}}^{k}_{t-1},{{\bf{c}}}^{q,k}_{t-1}])$.

Next, the new context vectors, i.e., ${\bf{c}}^{k}_t$ and
${\bf{c}}^{k,q}_{t}$, the attention distribution $\alpha^{k}_{t,i}$ for
each time step for the $k$-th passage, and the attention distribution
$\alpha^{k,q}_{t,i}$ for each time step for the question with the $k$-th
passage are computed as follows:

\vspace{-2mm}

\begin{eqnarray}
 {\bf{c}}^{k}_t \! \!\! \! &\!\!=\!\!& \! \!\! \!
 \sum_{i=1}^{N_k}{\alpha^{k}_{t,i}{\bf{u}}^{k}_i}, \quad {\bf{c}}^{k,q}_t \! \!
 = \! \! \sum_{i=1}^{N_q}{\alpha^{k,q}_{t,i}{\bf{h}}^{q}_{i}}, \quad \mbox{ s.t. }
\nonumber \\
 \alpha^{k}_{t,i} \! \!\! \! &\!\!=\!\!&\! \! \! \!
 \frac{\exp({e^{k}_{t,i}})}{\sum_{i=1}^{N_k}\exp(e^{k}_{t,i})}, \; \;
 e^{k}_{t,i} = \tanh(\!{\bf{w}}^k_h{\bf{u}}^{k}_i\! +\!
 {\bf{w}}^k_s{\bf{s}}^k_t \!+\! {{b}}^k_e), \nonumber \\
 \alpha^{k,q}_{t,i} \! \!\! \! &\!\!=\!\!&\! \! \! \!
 \frac{\exp({e^{k,q}_{t,i}})}{\sum_{i=1}^{N_q}\!\exp(e^{k,q}_{t,i})}, 
 e^{k,q}_{t,i} \! \!=\! \! \tanh(\!{\bf{w}}^{k,q}_{h}{\bf{h}}^{q}_{i}\! +\!
 {\bf{w}}^{k,q}_{s}{\bf{s}}^k_t \!+\! {{b}}^{k,q}_{e}). 
 \nonumber 
\end{eqnarray}
${\bf{w}}^k_h$, ${\bf{w}}^k_s$, ${\bf{w}}^{k,q}_{h}$, ${\bf{w}}^{k,q}_{s}$, ${{b}}^k_e$ and ${{b}}^{k,q}_{e}$ are learnable parameters.

Then, the output probability distribution over the vocabulary of words in the current state is computed for passage $k$:
\begin{equation}
 {\bf{V}}^k_{vocab} = \softmax({\bf{W}}^k[{\bf{s}}^k_t, {\bf{c}}^{k}_t, {\bf{c}}^{q,k}_{t}] + {\bf{b}}^k_v).\nonumber
\end{equation}
${\bf{W}}^k$ and ${\bf{b}}^k_v$ are learnable parameters. The number of rows in ${\bf{W}}^k$ represents the number of words in the vocabulary.

{\tabcolsep=1.7mm
\begin{table}[t]
\begin{center}
\caption{Ablation study of {\it GUM-MP} (MS-MARCO).}
 \footnotesize
{\tabcolsep = 1.0mm
\begin{tabular}{p{1.4cm}|p{1.1cm}p{1.1cm}p{1cm}p{1cm}p{1cm}}
Metric & {\it w/o Neg} & {\it w/o UM} & 
{\it UM(10)}& {\it UM(30)} & {\it UM(50)} \\ \hline

BLEU-1& 0.491 & 0.484 & 0.503 &0.501& {\bf{\em 0.514}}\\
ROUGE-L& 0.557 & 0.544 & {\bf{\em 0.569}} &0.568& 0.563 \\\hline
\end{tabular}
}
 \label{tab:resultMSAbl}
\end{center}
 \vspace{-3mm}
\end{table}

}
 \normalsize

{\tabcolsep=1.7mm
\begin{table}[t]
\begin{center}
\caption{Performance comparison (MS-MARCO).}
\footnotesize
{\tabcolsep = 1.0mm
\begin{tabular}{p{1.5cm}|p{1cm}p{1cm}p{1cm}p{1cm}p{1.3cm}}
Metric & {\it Trans} & {\it MPQG} & {\it S-Net} & 
{\it V-Net}& {\it GUM-MP} \\ \hline
BLEU-1& 0.060 & 0.342 & 0.364 & 0.407 & {\bf{\em 0.503}}\\
 ROUGE-L& 0.062 & 0.451 & 0.383 & 0.405 & {\bf{\em 0.569}} \\\hline

\end{tabular}
}
 \label{tab:resultMS}

\end{center}
 \vspace{-3mm}
\end{table}
}
 \normalsize

{\tabcolsep=1.7mm
\begin{table}[t]
\begin{center}
\caption{Ablation study of {\it GUM-MP} (Oshiete-goo).}
\footnotesize
{\tabcolsep = 1.0mm
\begin{tabular}{p{1.4cm}|p{1.1cm}p{1.1cm}p{1cm}p{1cm}p{1cm}p{1cm}}
Metric & {\it w/o Neg} & {\it w/o UM} & 
{\it UM(5)}& {\it UM(10)} & {\it UM(30)} \\ \hline

BLEU-1& 0.125 & 0.129 & 0.309 & 0.283& {\bf{\em 0.321}}\\
ROUGE-L& 0.224 & 0.222 & 0.253 & 0.248& {\bf{\em 0.265}} \\\hline
\end{tabular}
}
 \label{tab:resultOshiAbl}

\end{center}
 \vspace{-3mm}
\end{table}
}	
 \normalsize

{\tabcolsep=1.7mm
\begin{table}[t]
\begin{center}
\caption{Performance comparison (Oshiete-goo).}
 \footnotesize
{\tabcolsep = 1.0mm
\begin{tabular}{p{1.5cm}|p{1cm}p{1cm}p{1cm}p{1cm}p{1.3cm}}
Metric & {\it Trans} & {\it MPQG} & {\it S-Net} & 
{\it V-Net}& {\it GUM-MP} \\ \hline
BLEU-1& 0.041 & 0.232 & 0.247 & 0.246 & {\bf{\em 0.321}}\\
ROUGE-L& 0.088 & 0.251 & 0.244 & 0.249 & {\bf{\em 0.265}} \\\hline

\end{tabular}
}
 \label{tab:resultOshi}

\end{center}
 \vspace{-3mm}
\end{table}
}
 \normalsize

GUM-MP then utilizes the multiple-pointer-generator mechanism to compute
the final vocabulary distribution to determine the $t$-th answer word
$a_t$. It first computes the vocabulary distribution computed for
passage $k$ by interpolation between three probability distributions,
${\bf{V}}^k_{vocab}$, ${\bf{P}^k}_{attn}$, and
${\bf{Q}^k}_{attn}$. Here, ${\bf{P}^k}_{attn}$ and ${\bf{Q}^k}_{attn}$
are computed on the basis of the attention distributions
$\alpha^{k}_{t,i}$ and $\alpha^{k,q}_{t,i}$. It then integrates the
vocabulary distributions computed for each passage to compute the final
vocabulary distribution as follows:

\vspace{-2mm}
\begin{eqnarray}
{\bf{V}}_{final} \!\!&\!\!=\!\!&\!\! \sum_{k}^{K} (g_v^k {\bf{V}}^k_{vocab} + g^k_a
 {\bf{P}}^k_{attn} + g^k_q {\bf{Q}}^k_{attn}) \label{eq:final}
 \mbox{ s.t. }\quad \\ 
g^k_v, g^k_a, g^k_q \!\!&\!\!=\!\!&\!\!
\softmax({\bf{W}}^g
[{{\bf{s}}}^k_{t},{{\bf{c}}}^k_{t},{{\bf{c}}}^{q,k}_{t}]+{\bf{b}}_g)\nonumber
\end{eqnarray}
%where ${\bf{W}}^g \in {\mathbb{R}}^{3 \!\times\! 3d}$ and ${\bf{b}}_g \in {\mathbb{R}}^{3}$ are learnable parameters.
where ${\bf{W}}^g$ and ${\bf{b}}_g$ are learnable parameters.

Our multiple-pointer-generator naively checks the distributions
generated for each passage. With the UM, it determines which tokens in
the passages are important or not for answer generation. This improves
the generation accuracy.

\subsection{Training}
GUM-MP trains the model by optimizing the log-likelihood of the
gold-standard output sequence $\bf{A}$ with the cross-entropy loss
($\theta$ represents the trainable model parameters):

\begin{eqnarray}
{\cal{L}} = - {\sum_{t=1}^{N_a}
 \log p({\bf{a}}_t|\bf{Q},{\bf{a}}_1,\ldots,{\bf{a}}_{t-1};\theta)}. \nonumber 
\end{eqnarray}

\section{Evaluation}
This section evaluates GUM-MP in detail.

\subsection{Compared methods}
\label{sec:comparedMethods} We compared the performance of the following
five methods: (1) {\em Trans} is a Transformer
\cite{vaswani2017attention} that is used for answer generation. It
receives questions, not passages, as input; (2) {\em S-Net}
\cite{DBLP:conf/aaai/TanWYDLZ18}, (3) {\em MPQG}
\cite{DBLP:conf/naacl/SongWHZG18}, and (4) {\em V-Net}
\cite{wang-etal-2018-multi-passage}: these are explained in Section
\ref{sec:related}, though we applied our multiple-pointer-generator
mechanism to {\em MPQG} to make it handle multiple passages. (5) {\em
GUM-MP} is our proposal.

\subsection{Datasets}
We used the following two datasets:

\paragraph{MS-MARCO}
The questions are user queries issued to the Bing search engine, and
approximately ten passages and one answer are assigned to each
question. Among the official datasets in the MS-MARCO project, we chose
the natural language generation dataset, since our focus is answer
generation rather than answer extraction from passages. Human editors in
the MS-MARCO project reviewed the answers to the questions and rewrote
them as well-formed ones so that the answers would make sense even
without the context of the question or the retrieved passages. We
pre-trained a Glove model and also fine-tuned a publicly available
Bert-based model \cite{DBLP:journals/corr/abs-1810-04805} by using this
dataset. We then randomly extracted one-tenth of the full dataset
provided by the MS-MARCO project. The training set contained 16,500
questions and the test set contained 2,500 questions. The questions,
passages, and answers had on average 6, 68, and 16 words,
respectively.
%There are 3.8 million word tokens.

\paragraph{Oshiete-goo}
This dataset focused on the ``relationship advice'' category of the
Japanese QA community, Oshiete-goo \cite{AAAI20}. It has 771,956 answer
documents to 189,511 questions. We pre-trained the word embeddings by
using a Glove model on this dataset. We did not use the Bert model since
it does not improve the accuracy much. Then, human editors abstractly
summarized 24,661 answer documents assigned to 5,202 questions into
10,032 summarized answers\footnote{This summarized answer dataset is
used in the actual AI relationship advice service:
``https://oshiete.goo.ne.jp/ai''.}. Since the topics in several of the
answer documents assigned to the question overlap, the number of
summarized answers is smaller than the number of original answer
documents. Then, we randomly chose one-tenth of the questions as the
test dataset. The rest was used as the training dataset. The questions,
answer documents (hereafter, we call them passages), and summarized
answers (hereafter, we call them answers) had on average 270, 208, and
48 words, respectively. There were 58,955 word tokens and on average
4.72 passages for a question.

\subsection{Methodology and parameter setup}
\label{sec:methodology} To measure performance, we used BLEU-1
\cite{Papineni:2002:BMA:1073083.1073135} and ROUGE-L
\cite{lin:2004:ACLsummarization}, which are useful for measuring the
fluency of generated texts.

In testing the model, for both datasets, the negative passages were
selected from passages that were not assigned to the current
question. In the Oshiete-goo dataset, there were no passages assigned to
newly submitted questions. Thus, we used the answer selection method
\cite{TanSXZ16} to learn passage selection using the training
dataset. Then, we selected three positive passages per question from
among all the passages in the training dataset in the Oshiete-goo
dataset.
%This setting is natural for many applications.

%In training and testing
%the model, negative passages were randomly selected from XXX passages
%that were not assigned to the ``relationship advice'' category in
%Oshiete-goo.

We set the word embedding size to 300 and the batch size to 32. The
decoder vocabulary was restricted to 5,000 according to the frequency
for the MS-MARCO dataset. We did not restrict the attention
vocabularies. The decoder vocabulary was not restricted for the
Oshiete-goo dataset. Each question, passage, and answer were truncated
to 50, 130, and 50 words for the MS-MARCO dataset (300, 300, and 50
words for the Oshiete-goo one). The epoch count was $30$, the learning
rate was 0.0005, $Z$ in MPM was 5, and the beam size was 20.

\subsection{Results}
\label{sec:results}
\begin{table}[t]
 \begin{center}

\doublerulesep=1mm \caption{Example of answers output by the methods.} \footnotesize
 {\tabcolsep = 1.0mm
 \begin{tabular}{p{1.3cm}|p{6.8cm}} \hline
Question & Largest lake of USA? \\
\hline
  Passage 1 & The largest lake (by surface area) in the United States is {\bf lake michigan} with an area of 45410 square miles.
\\ \hline 
  Passage 2 & Iliamna lake is the largest lake in alaska and the second largest freshwater lake contained wholly within the United States (after {\bf lake michigan}).
 \\ \hline
Passage 3 & Superior is the largest lake that's partly in the United States at 31,820 square miles. The largest lake entirely contained in the United States is {\bf lake michigan}, 22,400 square miles.\\ \hline
{\em MPQG} & Lake is the largest lake of lake.\\ \hline
 {\em V-Net} & Is the largest lake that's partly in the United States at 31,820 square miles.\\ \hline
 {\em GUM-MP} & Lake michigan is the largest lake of the United States.\\ \hline
 Answer & The largest lake of United States of America is lake michigan.\\ \hline
 \end{tabular} }
 \normalsize
 \label{tab:MeanEx} \vspace{-4mm}\end{center}
 \end{table}

Table \ref{tab:resultMSAbl} and Table \ref{tab:resultOshiAbl} summarize
the ablation study for the MS-MARCO dataset and for Oshiete-goo
dataset. They compare several methods, each of which lacks one function
of {\em GUM-MP}; {\em w/o Neg} lacks the ``matching tensor'' in the
question and passage encoders. {\em w/o UM} lacks the ``Unified
Memories''. {\em UM(L)} is our method {\em GUM-MP} with $L$, the length
of row in PAM in Section \ref{sec:um-pam}.

The results indicate that all of the above functions are useful for
improving the accuracy of the generated answers for both datasets: {\em
GUM-MP} is better than {\em w/o Neg}. This means that the matching
tensor is useful for extracting good tokens from the answer or question
passages when generating the answers. {\em GUM-MP} is also superior to
{\em w/o UM}. This is because {\em GUM-MP} utilizes the UMs that include
the analysis of topics described in the passages for generating an
answer.
%
%Moreover, for the MS-MARCO dataset, the size of $L$ impacts the
%accuracy. We think this is because the suitable granularity of the
%topics described in the passages varies with the question, as the
%MS-MARCO dataset has very diverse QA pairs.
Especially, for the Oshiete-goo dataset, {\em GUM-MP} is much better
than {\em w/o UM} regardless of the size of $L$. This is because this
dataset focuses on ``relationship advice'' whereas the MS-MARCO dataset
has very diverse QA pairs; thus, the PAM well aligns the topics
described in multiple passages for the question in the Oshiete-goo
dataset.
%
%
%On the otherhand, for
%the MS-MARCO dataset, the suitable granularity of the topics described
%in the passages varies with the question, as the MS-MARCO dataset has
%very diverse QA pairs. 

Table \ref{tab:resultMS} and Table \ref{tab:resultOshi} summarize the
results of all of the compared methods on the MS-MARCO dataset and on
the Oshiete-goo dataset.
Here, we present {\em UM(10)} as {\em GUM-MP} in the MS-MARCO dataset
and {\em UM(30)} as {\em GUM-MP} in the Oshiete-goo dataset since they
are the best accurate results when changing $L$.
{\em S-Net} and {\em MPQG} generated better
results than {\em Trans} on both datasets, since they can make use of
passage information to extract tokens important for making the answers,
while {\em Trans} cannot. Looking at the generated answers, it is clear
that {\em MPQG} tends to extract tokens from multiple passages while
generating common phrases from the training vocabulary. In contrast,
{\em S-Net} tends to extract whole sentences from the passages. {\em
V-Net} tends to select good answer candidate spans from the passages;
however, it fails to generate answers that are edited or changed from
the sentences in the passages. Finally, {\em GUM-MP} is superior to {\em
V-Net} on both datasets, as it can identify the important word tokens
and at the same time avoid including redundant or noisy phrases across
the passages in the generated answers.

\subsection{Meaningful results}
Table \ref{tab:MeanEx} presents examples of answers output by {\em
MPQG}, {\em V-Net}, and {\em GUM-MP} on the MS-MARCO dataset. {\em MPQG}
mistakenly extracted a token since it could not consider the correlated
topics such as ``lake michigan'' within the passages. {\em V-Net}
extracted the exact text span taken from a passage that includes
redundant information for the question. {\em GUM-MP} points to the
tokens (e.g. lake michigan) that are described across multiple passages
that match the current question. As a result, it accurately generates
the answer.

\section{Conclusion}
\label{sec:conclusion} We proposed the neural answer Generation model
through Unified Memories over Multiple Passages (GUM-MP). GUM-MP uses
positive-negative passage analysis for passage understanding following
the question context. It also performs inter-relationship analysis among
multiple passages. It thus can identify which tokens in the passages are
truly important for generating an answer. Evaluations showed that GUM-MP
is consistently superior to state-of-the-art answer generation
methods. We will apply our ideas to a Transformer-based encoder-decoder
model.

\newpage

%\bibliographystyle{named}
%\bibliography{ijcai13}

\end{document}